\documentclass{article}
\usepackage{spconf,amsmath,graphicx,amssymb}
\usepackage{bbm}
\usepackage{algorithm}
\usepackage{algorithmic}
\usepackage{url}

\title{Superpixel Segmentation via Convolutional Neural Networks\\
with Regularized Information Maximization}

\name{Teppei Suzuki}
\address{Denso IT Laboratory, Inc.\\ 2-1-15 Shibuya, Shibuya-ku, Tokyo, Japan}
\begin{document}
%
\maketitle
\begin{abstract}
We propose an unsupervised superpixel segmentation method by optimizing a randomly-initialized convolutional neural network (CNN) in inference time. 
Our method generates superpixels via CNN from a single image without any labels by minimizing a proposed objective function for superpixel segmentation in inference time.
There are three advantages to our method compared with many of existing methods: (i) leverages an image prior of CNN for superpixel segmentation, (ii) adaptively changes the number of superpixels according to the given images, and (iii) controls the property of superpixels by adding an auxiliary cost to the objective function. 
We verify the advantages of our method quantitatively and qualitatively on BSDS500 and SBD dataset.\footnote{the code is available at \url{https://github.com/DensoITLab/ss-with-RIM}}
\end{abstract}
\section{Introduction}

Superpixels are a low-dimensional representation for images, and generally given as a set of pixels similar in color and other low-level properties (e.g., SLIC~\cite{slic}, SEEDS~\cite{seeds}, Felzenswalb and Huttenlocher's method (FH)~\cite{graph_sp}).
Superpixel segmentation is generally used as preprocessing for image processing tasks.

Many existing methods depend on local and low-level properties such as local connectivity, color, and positions to generate superpixels.
If the number of superpixels is large, the methods using the properties work well because the images almost consist of low-frequency components, and locally consist of pixels having the same color.
However, if the number of superpixels is small, superpixels need to group a wide range of pixels with various properties, and it is a difficult task for the methods using the local and low-level properties.


To introduce non-local properties into superpixel segmentation, we build a CNN-based superpixel segmentation method.
According to \cite{dip}, CNN has a prior for images, even though it is not trained.
Indeed, CNN can produce much cleaner results with sharper edges for inverse problems than methods using hand-crafted prior.
To leverage the prior, We utilize the same procedure as deep image prior (DIP)~\cite{dip}, which optimizes a randomly-initialized CNN using a single image in inference time without labels.
We assume that the prior also works well for superpixel segmentation, especially for capturing a global structure.
If the prior works as expected, our CNN-based method should have better performance than other methods with a small number of superpixels.


We design an objective function to execute superpixel segmentation with the DIP procedure.
The proposed objective function is inspired by regularized information maximization (RIM)~\cite{rim,mut_info}; we introduce a hyperparameter $\lambda$ to a mutual information term of RIM to control the number of superpixels.
Because of $\lambda$, our CNN-based method adaptively changes the number of superpixels depending on the given images.
Kanezaki~\cite{kanezaki} proposes an unsupervised segmentation method similar to ours, which optimizes the randomly-initialized CNN with a single image in inference time.
However, it does not focus on the superpixel segmentation, but rather the image segmentation so that it lacks a viewpoint that the number of segments is controllable.
The viewpoint is important for superpixel segmentation, because the required number of superpixels depends on subsequent tasks.
Thus, introducing $\lambda$ is a simple but key component for our method.


In our understanding, there are two types of superpixels, task-agnostic and task-specific superpixels, and we tackle the generation of the task-agnostic superpixels in this work.
Jampani \textit{et al.}~\cite{ssn} propose a supervised superpixel segmentation method, called superpixel sampling network (SSN), to generate the task-specific superpixels.
Unlike other superpixel segmentation methods, SSN is a supervised method that requires label information (e.g., semantic labels and optical flow labels) to train a model, and generates task-specific superpixels.
If the subsequent processing requirements are known, SSN is an effective method.
However, it is difficult for some tasks, such as optical flow and depth estimation in natural scenes, to define and annotate ground-truth labels.
Therefore, unsupervised superpixel segmentation is a crucial task.
In terms of unsupervised setting, superpixels should retain the information of the original images to be able to respond to any tasks.
We refer to the superpixels satisfying it as the task-agnostic superpixels, and to generate them, we introduce the reconstruction cost as an auxiliary cost to our CNN-based method.


In experiments, we verify the effectiveness of our method quantitatively and qualitatively.
With a quantitative evaluation, we study the relationship between $\lambda$ and the number of superpixels, and compare our method to baseline methods~\cite{slic,seeds,graph_sp,etps} on BSDS500~\cite{bsd} and SBD~\cite{sbd} datasets.
With a qualitative evaluation, we verify that the image reconstruction as an auxiliary task brings a positive effect to retain the information of the original image by comparing example results.

\section{Superpixel Segmentation via Convolutional Neural Networks}
\subsection{Preliminary}
Let $I\in\mathbb{R}^{H\times W\times C}$ be an input image, where $H$, $W$, and $C$ denote image height, width, and input channels (typically RGB), respectively.
Our goal is to assign superpixels $\mathcal{S}=\{s_1,\dots,s_N\}$ to all pixels, where $s_n$ denotes $n$-th superpixel to be a set of pixels, and the number of superpixels $N$ is a hyperparameter. Note that our CNN-based method allows $s_n$ to be an empty set. Therefore, the hyperparameter $N$ is the upper bound of the number of superpixels.

We define superpixel segmentation as $N$-class classification problem.
Let $P\in\mathbb{R}^{H\times W\times N}_+;\:\sum_nP_{h,w,n}=1$ be probabilistic representation of superpixels, where $\mathbb{R}_+$ indicates non-negative real number. The superpixel assigned to a pixel at $(h, w)$ is given by $\mathrm{arg}\max_nP_{h,w,n}$.
We obtain $P$ through optimization of an objective function defined in the next section.

\subsection{Framework}
We define an objective function for our CNN-based superpixel segmentation method using the DIP procedure.
The pseudocode of our method is shown in Algorithm 1.

The objective function consists of three parts as follows:
\begin{align}
    \mathcal{L}_\mathrm{objective} = \mathcal{L}_\mathrm{clustering}+\alpha\mathcal{L}_\mathrm{smoothness} + \beta\mathcal{R},
\end{align}
where $\mathcal{L}_\mathrm{clustering}$, $\mathcal{L}_\mathrm{smoothness}$, and $\mathcal{R}$ denote an entropy-based clustering cost, spatial smoothness, and an additional term that is the reconstruction cost in this work, respectively.
$\alpha$ and $\beta$ are hyperparameters balancing the importance of each term. 
As shown in Algorithm 1, by minimizing $\mathcal{L}_\mathrm{objective}$ with respect to parameters of CNN, we obtain $\mathcal{S}$.

\begin{algorithm}[t]
\caption{CNN-based superpixel segmentation}
\label{alg:cnnsp}
\begin{algorithmic}[1]
\STATE \textbf{Input:} An image $I\in\mathbb{R}^{H\times W\times C}$; pixel locations $X\in\mathbb{R}^{H\times W\times 2}$; hyperprameters: the number of superpixels $N$, coefficients $(\lambda,\alpha,\beta)$, number of iteration $T$, and learning rate $\eta$
\STATE \textbf{Output:} Superpixels $\mathcal{S}$
\STATE Initialize CNN with randomly-sampled parameters $\theta$, \\$f_\theta:(I,X)\rightarrow (P,\hat{I})$
\FOR{$t=1,\dots, T$}
\STATE Get probability and reconstructed image, \\$(P,\hat{I})\leftarrow f_\theta(I,X)$
\STATE Calculate $\mathcal{L}_\mathrm{objective}$
\STATE Update parameters by gradient descent, \\$\theta\leftarrow\theta-\eta\frac{\partial}{\partial\theta}\mathcal{L}_\mathrm{objective}$
\ENDFOR
\STATE Assign superpixels to all pixels, $\mathrm{arg}\max_n P_{h,w,n}$

\end{algorithmic}
\end{algorithm}

$\mathcal{L}_\mathrm{clustering}$ is an entropy-based clustering cost, which is similar to a mutual information term of regularized information maximization (RIM)~\cite{rim,mut_info}. $\mathcal{L}_\mathrm{clustering}$ is as follows:
\begin{align}
\nonumber
    \mathcal{L}_\mathrm{clustering}=&\frac{1}{HW}\sum_{h,w}\sum_n-P_{h,w,n}\log P_{h,w,n}\\
    &+\lambda\sum_n\hat{P}_n\log\hat{P}_n,
\end{align}
where $\hat{P}\in\mathbb{R}^N$ denotes the mean value of the probability vectors over all pixels, $\hat{P}_n=\frac{1}{HW}\sum_{h,w}P_{h,w,n}$.
The first term is the entropy of $P_{h,w}\in\mathbb{R}^N$, and minimization of it encourages deterministic superpixel assignment.
The second term is the negative entropy of the mean vector over all probability vectors, and minimization of it encourages the size of each superpixel to be uniform.

Unlike the mutual information term of RIM, we introduce a scalar value $\lambda$ as a coefficient of the second term to control the number of superpixels.
When $\lambda$ is small, the model tends to try to segment an image with a small number of superpixels because the first term becomes dominant, and the model ignores the second term.
As $\lambda$ increases, the number of superpixels given by CNN converges on $N$.
However, if $\lambda$ is too large, $P_{h,w}$ tends to become uniform because the second term becomes dominant.
In practice, our method works well when $\lambda$ is within $[0,3]$.
In experiments, we study the effect of $\lambda$ in detail.

The smoothness term is a primary prior for image processing tasks, which quantifies the difference between adjacent pixels.
We define $\mathcal{L}_\mathrm{smoothness}$ as follows:
\begin{align}
    \nonumber
    \mathcal{L}_\mathrm{smoothness}=\frac{1}{HW}\sum_{h,w}\left(\left\|\partial_xP_{h,w}\right\|_1e^{-\|\partial_x I_{h,w}\|_2^2/\sigma}\right.\\
    +\left.\left\|\partial_yP_{h,w}\right\|_1e^{-\|\partial_y I_{h,w}\|_2^2/\sigma}\right),
\end{align}
where $\partial$ and $\|\cdot\|_p$ denote the image gradients and $p$-norm.
$P_{h,w}\in\mathbb{R}^N_+$ and $I_{h,w}\in\mathbb{R}^C$ are the vectors at $(h,w)$ of $P$ and $I$, respectively.
$\sigma$ is a scalar value, and we set it to 8 in experiments.
$\mathcal{L}_\mathrm{smoothness}$ is the same as proposed in \cite{monodepth}.


In this work, we provide a reconstruction cost $\mathcal{R}_\text{recons}$ as an auxiliary task.
We consider that a single CNN outputs both the probabilities $P$ and the reconstructed image $\hat{I}\in\mathbb{R}^{H\times W\times C}$.
The reconstruction cost is defined as follows:
\begin{align}
    \mathcal{R}_\text{recons}=\frac{1}{HWC}\sum_{h,w}\|I_{h,w}-\hat{I}_{h,w}\|_2^2.
\end{align}
This is a primary loss function for autoencoders.
To minimize $\mathcal{R}_\text{recons}$, CNN needs to retain structures of the image in intermediate representations.
Therefore, we expect that adding $\mathcal{R}_\text{recons}$ to the objective function fit superpixels to detail components in images.
In experiments, We verify the reconstruction cost works as expected.

\begin{figure}[t]
    \centering
    \includegraphics[clip,width=1\hsize]{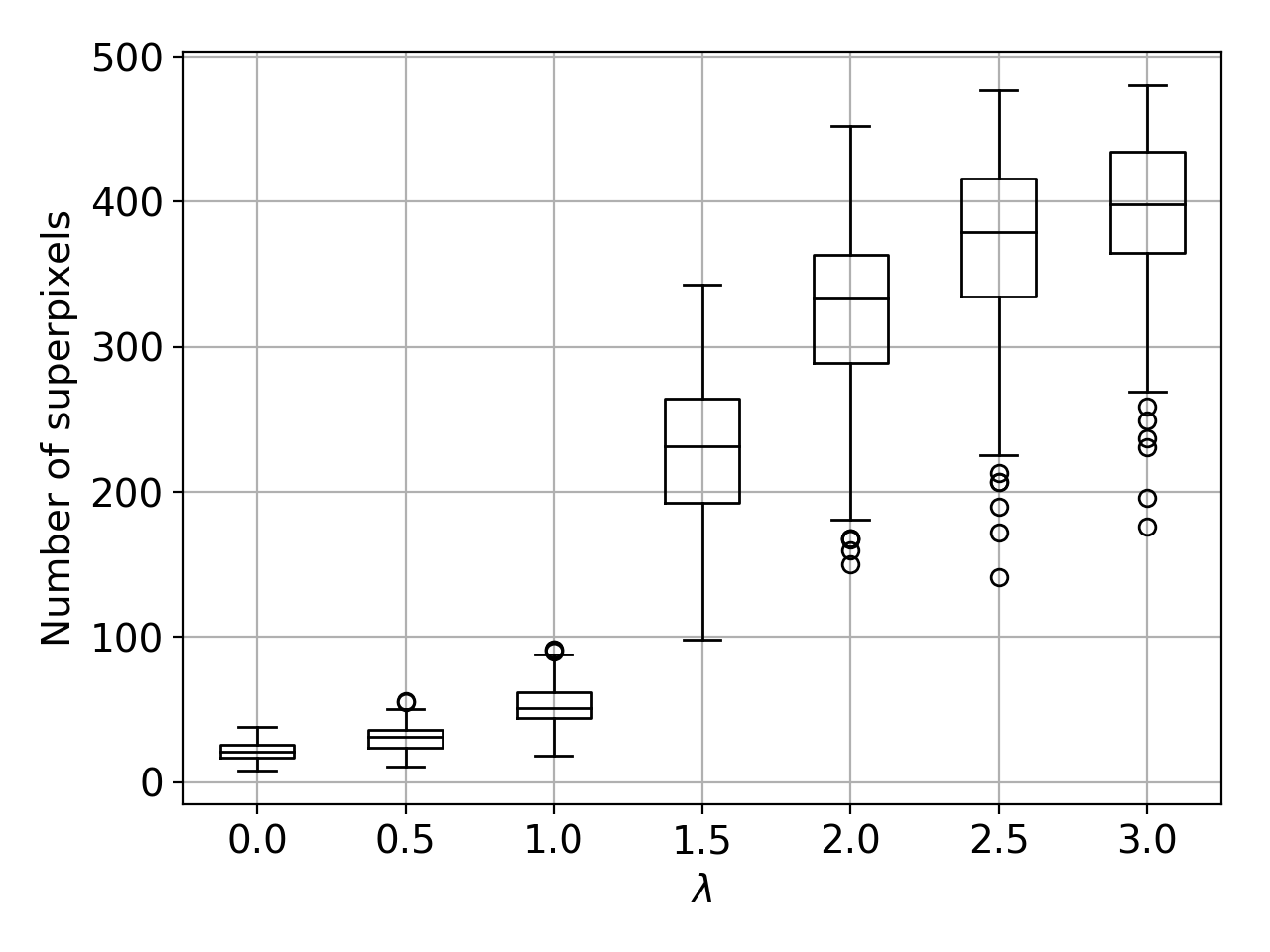}
    \caption{The number of superpixels per image on the BSDS500~\cite{bsd} test image set with various $\lambda$. We generate these results by the model optimized in eq. (1) with the reconstruction cost. The maximum number of superpixels $N$ is 500. The number of superpixels converges on a small number when $\lambda$ is small. On the other hand, when $\lambda$ is large, the number of superpixels spreads over a wide range, and the mean number of superpixels becomes large.}
    \label{fig:hitogram}
\end{figure}

\begin{figure*}[t]
    \centering
    \includegraphics[clip, width=1\hsize]{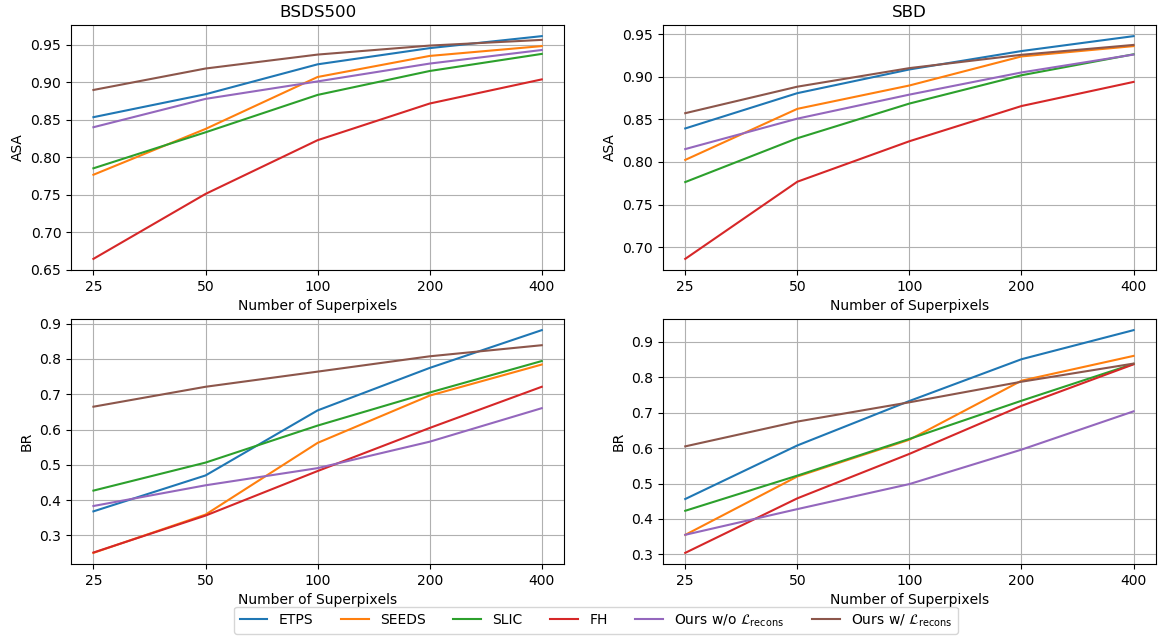}
    \caption{Comparison of proposed method to baseline methods~\cite{slic,seeds,graph_sp,etps}. We show achievable segmentation accuracy (ASA), and boundary recall (BR) on BSDS500~\cite{bsd} and SBD~\cite{sbd} with various numbers of superpixels. Ours w/ recons and w/o recons denote that the reconstruction cost is used for the optimization or not.}
    \label{fig:results_asa_rec}
\end{figure*}

\begin{figure*}[t]
    \centering
    \begin{tabular}{cccccc}
        \begin{minipage}{0.165\hsize}
        \centering
        \includegraphics[width=1\hsize]{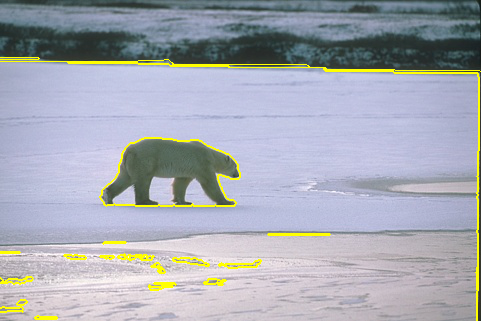}
        \end{minipage}
        \begin{minipage}{0.165\hsize}
        \centering
        \includegraphics[width=1\hsize]{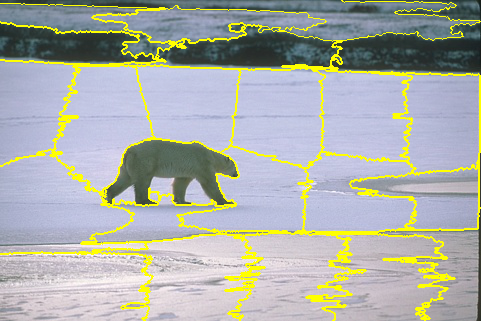}
        \end{minipage}
        \begin{minipage}{0.165\hsize}
        \centering
        \includegraphics[width=1\hsize]{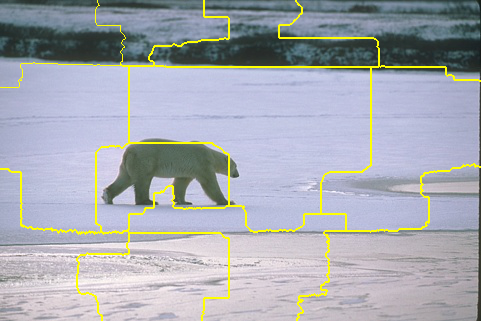}
        \end{minipage}
        \begin{minipage}{0.165\hsize}
        \centering
        \includegraphics[width=1\hsize]{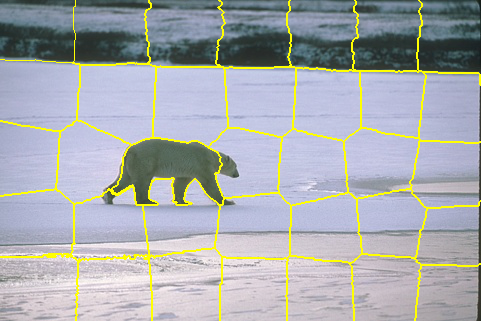}
        \end{minipage}
        \begin{minipage}{0.165\hsize}
        \centering
        \includegraphics[width=1\hsize]{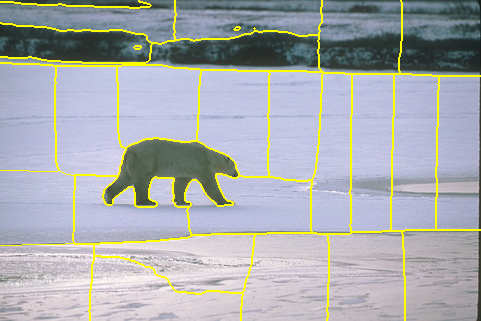}
        \end{minipage}
        \begin{minipage}{0.165\hsize}
        \centering
        \includegraphics[width=1\hsize]{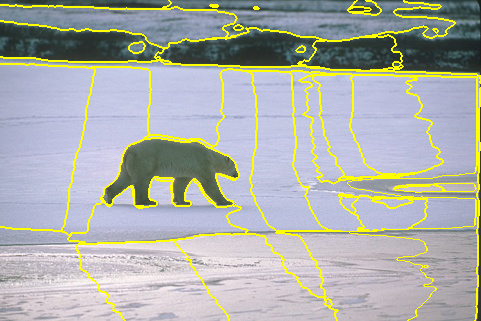}
        \end{minipage}\\
        \begin{minipage}{0.165\hsize}
        \centering
        \includegraphics[width=1\hsize]{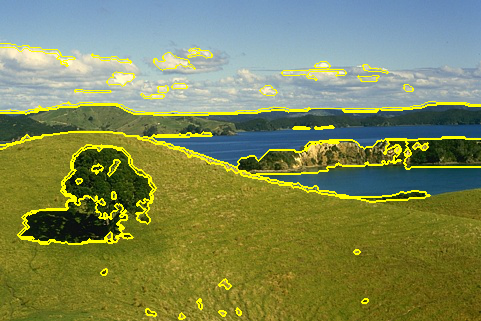}
        \end{minipage}
        \begin{minipage}{0.165\hsize}
        \centering
        \includegraphics[width=1\hsize]{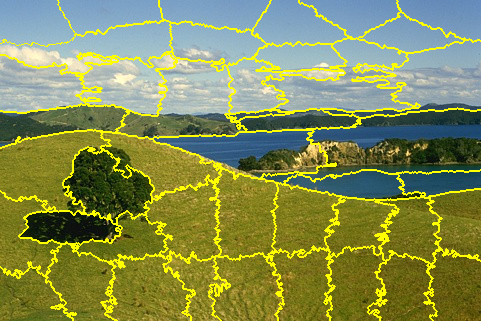}
        \end{minipage}
        \begin{minipage}{0.165\hsize}
        \centering
        \includegraphics[width=1\hsize]{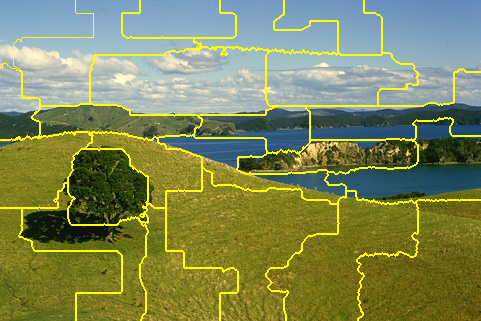}
        \end{minipage}
        \begin{minipage}{0.165\hsize}
        \centering
        \includegraphics[width=1\hsize]{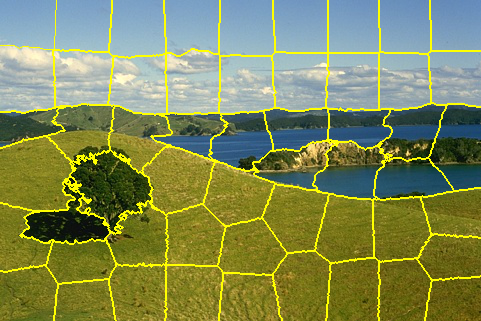}
        \end{minipage}
        \begin{minipage}{0.165\hsize}
        \centering
        \includegraphics[width=1\hsize]{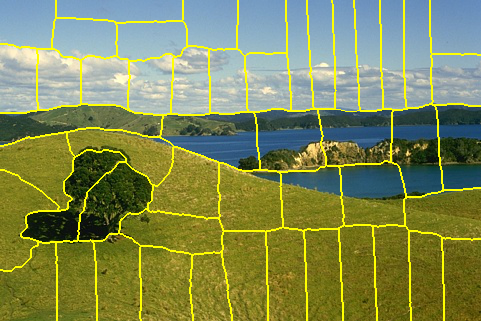}
        \end{minipage}
        \begin{minipage}{0.165\hsize}
        \centering
        \includegraphics[width=1\hsize]{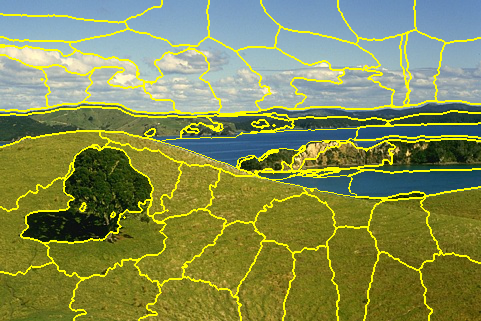}
        \end{minipage}\\
        \begin{minipage}{0.165\hsize}
        \centering
        \includegraphics[width=1\hsize]{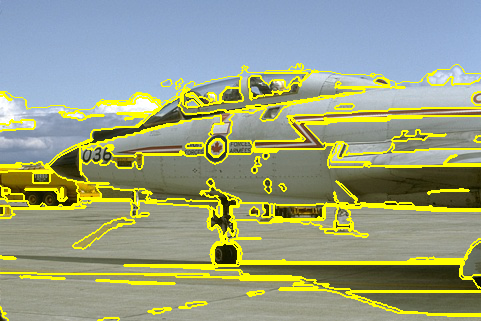}
        (a)
        \end{minipage}
        \begin{minipage}{0.165\hsize}
        \centering
        \includegraphics[width=1\hsize]{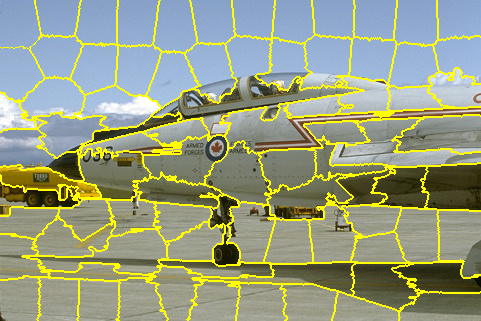}
        (b)
        \end{minipage}
        \begin{minipage}{0.165\hsize}
        \centering
        \includegraphics[width=1\hsize]{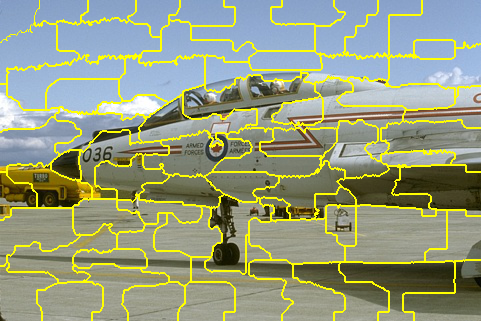}
        (c)
        \end{minipage}
        \begin{minipage}{0.165\hsize}
        \centering
        \includegraphics[width=1\hsize]{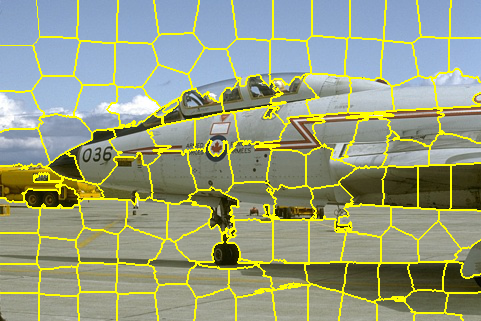}
        (d)
        \end{minipage}
        \begin{minipage}{0.165\hsize}
        \centering
        \includegraphics[width=1\hsize]{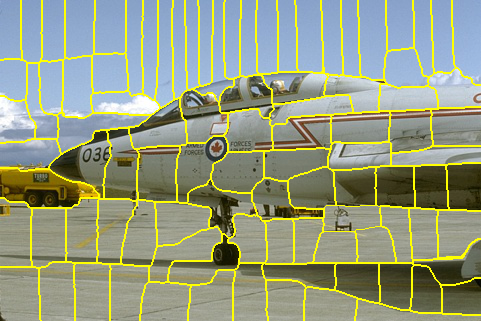}
        (e)
        \end{minipage}
        \begin{minipage}{0.165\hsize}
        \centering
        \includegraphics[width=1\hsize]{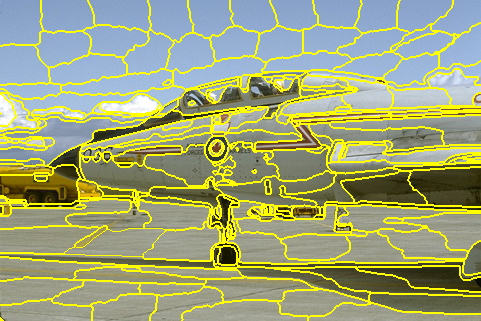}
        (f)
        \end{minipage}
    \end{tabular}
    \caption{Example results of (a) FH~\cite{graph_sp} (b) SLIC~\cite{slic} (c) SEEDS~\cite{seeds} (d) ETPS~\cite{etps} (e) ours without the reconstruction const, and (f) ours with the reconstruction cost. From top to bottom, the number of superpixels is 25, 50, and 100.}
    \label{fig:results_images}
\end{figure*}

\section{experiments}
To evaluate our framework, we study the effect of the coefficient of eq. (2), $\lambda$, and compare our method to the clustering-based method~\cite{slic}, the energy optimization~\cite{seeds,etps}, and the graph-based method~\cite{graph_sp} on the Berkeley Segmentation Dataset and Benchmarks (BSDS500)~\cite{bsd} and Stanford Background Dataset (SBD)~\cite{sbd}.
BSDS500 contains 300 train/validation images and 200 test images, and we use 200 test images for the evaluation in the experiments, and SBD contains 715 images, and we use all images for the evaluation.
We use implementations of OpenCV~\cite{opencv}, scikit-image~\cite{skimage}, and author's implementation~\footnote{https://bitbucket.org/mboben/spixel/src/master/} of ETPS for baseline methods, and utilize PyTorch~\cite{pytorch} to implement our method.

We use standard metrics for superpixel segmentation to evaluate the performance, achievable segmentation accuracy (ASA) and boundary recall (BR). 
ASA is a metric to quantify the achievable accuracy of superpixel-based segmentation.
It is defined as follows:
\begin{align}
    ASA(\mathcal{S},\mathcal{G})=\frac{\sum_i\max_{j}|s_i\cap g_j|}{\sum_i|g_i|},
\end{align}
where $\mathcal{G}=\{g_1,\dots,g_M\}$ denotes a set of ground-truth segments, and $g_m$ denotes a set of pixels.
BR quantifies recall of the boundary between segments in ground-truth labels. 
BR is defined as follows:
\begin{align}
    BR(\mathcal{B}^\mathcal{S},\mathcal{B}^\mathcal{G})=\frac{\mathrm{TP}(\mathcal{B}^\mathcal{G},\mathcal{B}^\mathcal{S})}{\mathrm{TP}(\mathcal{B}^\mathcal{G},\mathcal{B}^\mathcal{S})+\mathrm{FN}(\mathcal{B}^\mathcal{G},\mathcal{B}^\mathcal{S})},
\end{align}
where $\mathcal{B}^\mathcal{S}=\{b_1^\mathcal{S},\dots\}$ and $\mathcal{B}^\mathcal{G}=\{b_1^\mathcal{G}\dots\}$ denote a set of boundary pixels in $\mathcal{S}$ and $\mathcal{G}$.
FN and TP are the number of false negatives and true positives boundary pixels, respectively.
If a boundary pixel in $\mathcal{S}$ exists within a $(2\epsilon+1)\times(2\epsilon+1)$ local patch centered on an arbitrary boundary pixel in $\mathcal{G}$, the pixel is counted as TP. 
We set $\epsilon$ to $1$ in our experiments.

\subsection{Implementation details}
We evaluate our method with five-layer CNN with ReLU non-linearlity~\cite{relu}.
The channels for each layer except for the output layer are set to $32\cdot2^{l-1}$, where $l$ indicates the layer index, $l\in[0,5]$.
We use softmax activation to ensure $\sum_nP_{h,w,n}=1$, and apply instance normalization~\cite{inorm} for the feature map before softmax activation.
Use of instance normalization is suggested in \cite{kanezaki}. 
We optimize the model for 1,000 iterations by Adam~\cite{adam}.
We set $0.01$ to learning rate, and other parameters are the same as the default parameters.
The coefficients $(\lambda, \alpha, \beta)$ are set to $(2,2,10)$ in our experiments, but $\beta$ is zero for the model without the reconstruction cost.
These parameters were roughly selected by evaluation on the train/validation data.

In practice, if given only RGB image as input, CNN groups independent connected components as the same superpixel, because CNN has translation invariant, and assigns the superpixels based on only local spatial patterns.
Therefore, we also give pixel locations $X\in\mathbb{R}^{H\times W\times 2}$ to the model as input, namely $f:(I,X)\rightarrow(P,\hat{I})$, to reduce undesired segments. 
The pixel locations cannot completely prevent the undesired segments, but practically works well. Therefore, for implementation of the (RGB, location)-input (probability, reconstruction)-output model, we set input channels of CNN to 5, and output channels of CNN to $N+3$.
The inputs are normalized for each channel so that the mean and the variance of each channel of $X$ and $I$ are 0 and 1.

\subsection{Results}
We show the number of superpixels per image with various $\lambda$ in Fig.~\ref{fig:hitogram}.
We set $N$ to 500.
The model assigns various numbers of superpixels in each image, and the number of superpixels spreads over a wide range when $\lambda$ is large.
This indicates that our CNN-based method adapts the number of superpixels depending on the given images.
If one desires the superpixels to be as few as possible, $\lambda$ should be set a small value.
On the other hand, if one desires to adaptively control the number of superpixels, $\lambda$ should be set a large value.

Fig.~\ref{fig:results_asa_rec} shows that our method achieves comparable or better results compared to other methods in ASA with $\leq200$ superpixels.
As we expected, our methods clearly improve ASA in a small number of superpixels.
Our methods also achieves comparable or better BR with the small number of superpixels.
Especially, our method with the reconstruction dramatically improves BR with $\leq 200$ superpixels on BSDS500.
It indicates that the reconstruction cost refines the segmentation accuracy around the object boundaries.

We show example results of each method in Fig.~\ref{fig:results_images}.
The superpixels generated by our methods have properties between SLIC~\cite{slic} and FH~\cite{graph_sp}, especially ours with the reconstruction cost.
Ours with the reconstruction partially fit the superpixels to detail components in the images, such as the vertical tail of the helicopter and the body paints of the plane.
FH also seems to fit segments to detail components; however, its ASA is lower than other methods, and the fact indicates that FH groups pixels belonging to different segments in the ground-truth label.
It indicates that the segments generated by FH cannot preserve the semantic information of the input image.
Our methods achieve higher ASA, and qualitatively capture the detail components, preserving the information of the original image.

\section{conclusion}
We built the CNN-based superpixel segmentation method and proposed the objective function for superpixel segmentation.
We verified three advantages qualitatively and quantitatively: leveraging the prior of CNN, adjusting the number of superpixels depending on images, and retaining the information of the original image by the reconstruction cost.

Our method has two limitations: generated superpixels depend on the initial parameters of CNN, and independent connected components may belong to the same superpixel, especially when the method uses the reconstruction loss.
We believe that these limitations provide interesting research directions, one is ensemble modeling, and the other is topological data analysis.
We will explore the directions in future work.


\bibliographystyle{IEEEbib}
\bibliography{strings,refs}

\end{document}